\documentclass[runningheads]{llncs}
\usepackage{graphicx}
\usepackage[dvipsnames]{xcolor}
\usepackage{amsmath, amssymb}
\usepackage{url}

\begin{document}
\title{Blockchain as an Enabler for Transfer Learning in Smart Environments}

\author{Amin Anjomshoaa\orcidID{0000-0001-6277-742X} \and
Edward Curry\orcidID{0000-0001-8236-6433}}

\authorrunning{A. Anjomshoaa et al.}

\institute{
Lero – the Irish Software Research Centre ,\\ National University of Ireland, 
\email{amin.anjomshoaa,edward.curry}@mu.ie
}

\maketitle              
\begin{abstract}

The knowledge, embodied in machine learning models for intelligent systems, is commonly associated with time-consuming and costly processes such as large-scale data collection, data labelling, network training, and fine-tuning of models. Sharing and reuse of these elaborated models between intelligent systems deployed in a different environment, which is known as transfer learning, would facilitate the adoption of services for the users and accelerates the uptake of intelligent systems in environments such as smart building and smart city applications. In this context, the communication and knowledge exchange between AI-enabled environments depend on complicated networks of systems, system of systems, digital assets, and their chain of dependencies that hardly follows the centralized schema of traditional information systems. Rather, it requires an adaptive decentralized system architecture that is empowered by features such as data provenance, workflow transparency, and validation of process participants. In this research, we propose a decentralized and adaptive software framework based on blockchain and knowledge graph technologies that supports the knowledge exchange and interoperability between IoT-enabled environments, in a transparent and trustworthy way.

\end{abstract}
\section{Introduction}

Internet of Things (IoT) and cognitive computing are the two major enabling technologies in smart environment solutions such as smart home, smart building, and smart office use cases. Thanks to the proliferation of networking and communication trends and since the emergence of IoT, deploying the IoT infrastructure for sensing and monitoring of spaces is rather a straightforward practice. However, cognitive knowledge cannot be explicitly defined and is acquired by observation and experimentation. An example of this type of cognitive knowledge is machine learning models that are taught to undertake complex tasks and to create explicit and actionable knowledge from descriptive knowledge. 

The IoT-enabled systems are typically built following a three-layer model that consists of: (i) a sensing layer, which acquires the observation of interest from the environment; (ii) a cognitive layer, which is concerned with context acquisition, modeling, and reasoning; and (iii) an actuate layer that triggers an action or invokes a service according to some predefined logic. The main effort required for automatic change management in a private IoT space is hidden in the cognitive layer. Unlike the sense and actuate layers which typically undertake straightforward tasks, the creation and configuration of the context layer is a complicated task which is not necessarily based on a syntactic or deterministic model. Instead, it needs a deep understanding of incoming events as well as the context of those events. Currently, human intervention is necessary to interpret the incoming data from the sensor layer and add the missing semantics about the events and their context.

\begin{figure}
\centering
\includegraphics[width=1\textwidth]{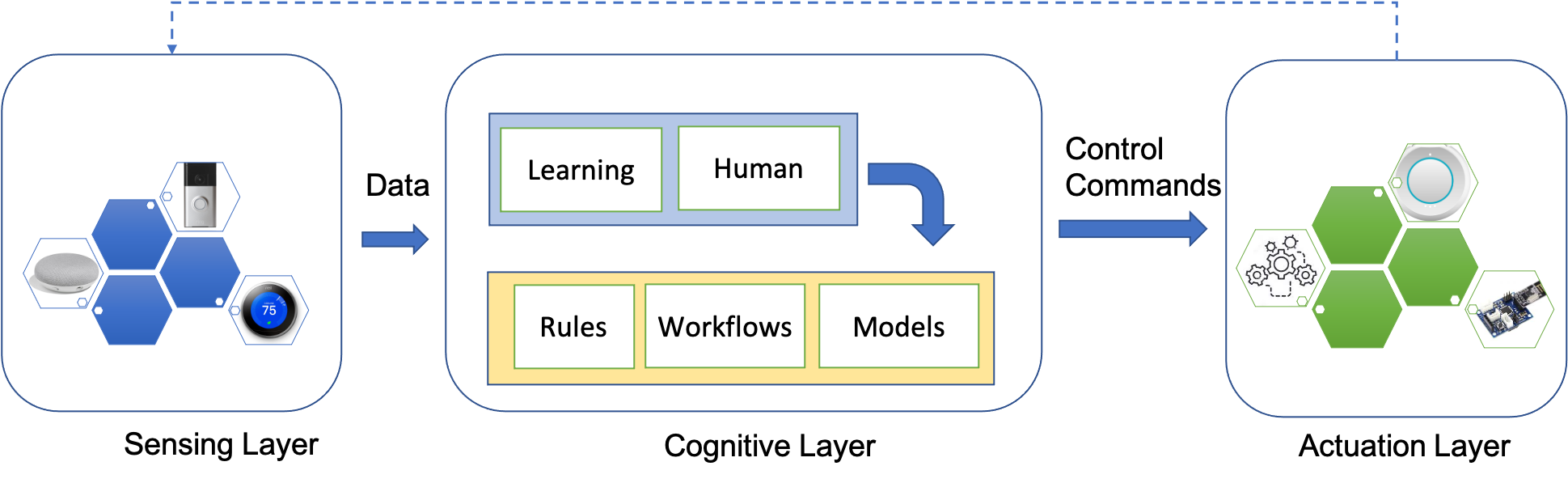}
\caption{Three-layer model of an IoT-based system including sensing, context, and actuation layers.}
\label{sense_actuate}
\end{figure}

In order to benefit the services of our IoT-enabled environments, we need to calibrate and configure them based on space-specific properties and semantics. These personalized configurations are commonly captured in dedicated information resources such as service compositions, sense-actuate cycles, and well-trained machine-learning models. In this context, the smart space services, embodied in machine learning models, are commonly associated with time-consuming and costly processes such as large-scale data collection, data labeling, network training, and fine-tuning models. Increasing access to the cognitive knowledge held by these models unlocks the value of transfer learning in smart environment and enables sustainable and autonomous growth of IoT-enabled environment. 

In the context of smart built environments, machine learning methods benefit from the huge amount of data in IoT-enabled spaces in their training processes. Such models could yield significant profit over the lifetime of a building but due to the high cost of training and the need of large amounts of training and testing data sets, the uptake of machine learning applications in built environments is rather slow. For instance, in a smart environment a speech recognition model could be trained based on a rather big and elaborated dataset. However, the model might not be used in different spaces as it should be retrained to cope with the surrounding noise of the target space. This gives smart environments an incentive to reuse and re-purpose the existing models via techniques such as transfer learning where models trained on one task are re-purposed for a second related task. 

Currently, machine learning processes are commonly built for intra-organization purposes and tailored towards specific use cases with the assumption of integrated model repositories and feature pools. In order to unleash the power of these resources, they need to be shared and reused in a trustworthy and transparent way which requires an adaptive decentralized system architecture that is empowered by features such as data provenance, workflow transparency, and validation of process participants. 

In order to make the machine learning processes for intelligent systems and their digital assets the first-class citizens of a decentralized and peer-to-peer ecosystem, the following challenges should be addressed: 

\begin{enumerate}

\item \textbf{Integrity:} As we move beyond the borders of ML model’s source environments (e.g. space, building, or city), we would need to fortify the integrity and traceability of adopted digital assets such as training datasets or transferred models.

\item \textbf{Transparency and trust:} in a decentralized process, the transparency and trustworthiness of actors and digital assets are of great importance. Although the transparency and trustworthiness are the main drivers behind the shift to blockchain technology, but issues such as communication between private and public blockchains and inter-organizational trust networks needs to be investigated via a holistic approach that takes the specific characteristics of machine learning pipelines into account. 
\item \textbf{Explainable machine learnig chains:} the transfer learning processes are commonly undertaken by a human who is able to understand the implicit semantics of environments, features, as well as the relevant information resources and services. Therefore, self-explaining and machine-understandable digital assets (e.g. features and models) are the key requirements to enable transfer learning beyond the organizational borders.

\end{enumerate}

In this research, we propose a decentralized and adaptive software framework based on blockchain and knowledge graph technologies for reliable and decentralized transfer of machine learning digital assets between AI-enabled smart environments such as smart buildings and smart city, in a transparent and trustworthy way. The proposed approach in this research follows the principles of Inter-Space Learning (ISL) framework \cite{anjomshoaa2021transfer} that was presented ion our previous work.  Using ISL framework, smart environment resources are processed, aligned, and reconciled to create a uniform and interconnected information space that facilitates provisioning of the Inter-space learning. This research an adaptive intelligent system architecture and advances the ISL framework by: 

\begin{enumerate}
\item Establishing a decentralized feature space that provides context for machine learning processes and makes their digital assets explainable and interoperable for transfer learning purposes. This will be achieved by creating and maintaining a knowledge graph that is aware of both machine learning features, their semantics, and the models trained based on those features. The knowledge graph includes also the evaluation indicators of machine learning models which facilitate the sharing and reuse of models beyond the organization boundaries. 
\item Integration of blockchains into machine learning pipelines in order to safeguard the data integrity and trustworthiness in cross-organization transfer learning scenarios. To this end, we exploit blockchain technology to capture the interaction between machine learning digital assets including raw data, features, and models. 
\item Creating a collaborative and adoptive framework based on public and private blockchains where participants can continually share, reuse, and evaluate machine learning assets in a reliable and incentive-oriented manner. As such, the participants will be able to benefit from publicly-traceable ML pipelines and integrate them in their secure intra-organization processes.
\end{enumerate}

\subsection{Building Industry as a Motivational Use Case}

Improvement of the energy performance of buildings is a major contributor for achieving the goals of the EU 2030 climate and energy framework \footnote{\url{https://ec.europa.eu/energy/sites/ener/files/documents/1.final_report.pdf}}. By 2050 at least 75\% of today’s buildings will still exist and as a result, renovation of existing buildings and making them energy-efficient will have significant economic and environmental impacts. The majority of legacy buildings were mainly built without considering the energy performance requirements and there is a high potential of saving in operational energy costs. Furthermore, by renovating the legacy buildings, the embodied energy of new constructions which makes up a major part of the total energy use in buildings, could be avoided.

Current renovation processes commonly extend the existing buildings with additional intelligence in order to optimize the operational costs. In this context, the major enabling technologies are IoT and cognitive computing. While deploying IoT infrastructure during the renovation processes is rather a straightforward practice, the implementation of cognitive components that benefit from generated data of IoT infrastructure is a complicated task that should be elaborated based on the specific attributes of tasks and buildings.

In smart buildings, such cognitive components can be realized via machine learning models that are taught to undertake complex tasks. The high costs of creating such models which require large amounts of training and testing datasets, gives smart environments the incentive to share and reuse the existing models. This research aims to investigate the potential of communication and reuse of cognitive knowledge between smart environments using blockchain technology.

\section{Knowledge Graph}
In our previous work, we introduced the Inter-Space Learning Framework (ISL) \cite{anjomshoaa2021transfer} Which uses knowledge graph as an efficient medium for communicating the semantics of smart environments and bridging the information gap among IoT infrastructure, spaces, and machine learning processes. 

As such the digital assets of machine learning processes are presented in an explainable and interoperable way which will facilitate the transfer of cognitive components. The proposed knowledge graph in ISL is aware of IoT infrastructure of target environments as well as machine learning features, their semantics, and the models trained based on those features. The knowledge graph includes also the evaluation indicators of machine learning models which facilitate the sharing and reuse of models beyond the organization boundaries. 

\begin{figure}
\centering
\includegraphics[width=1\textwidth]{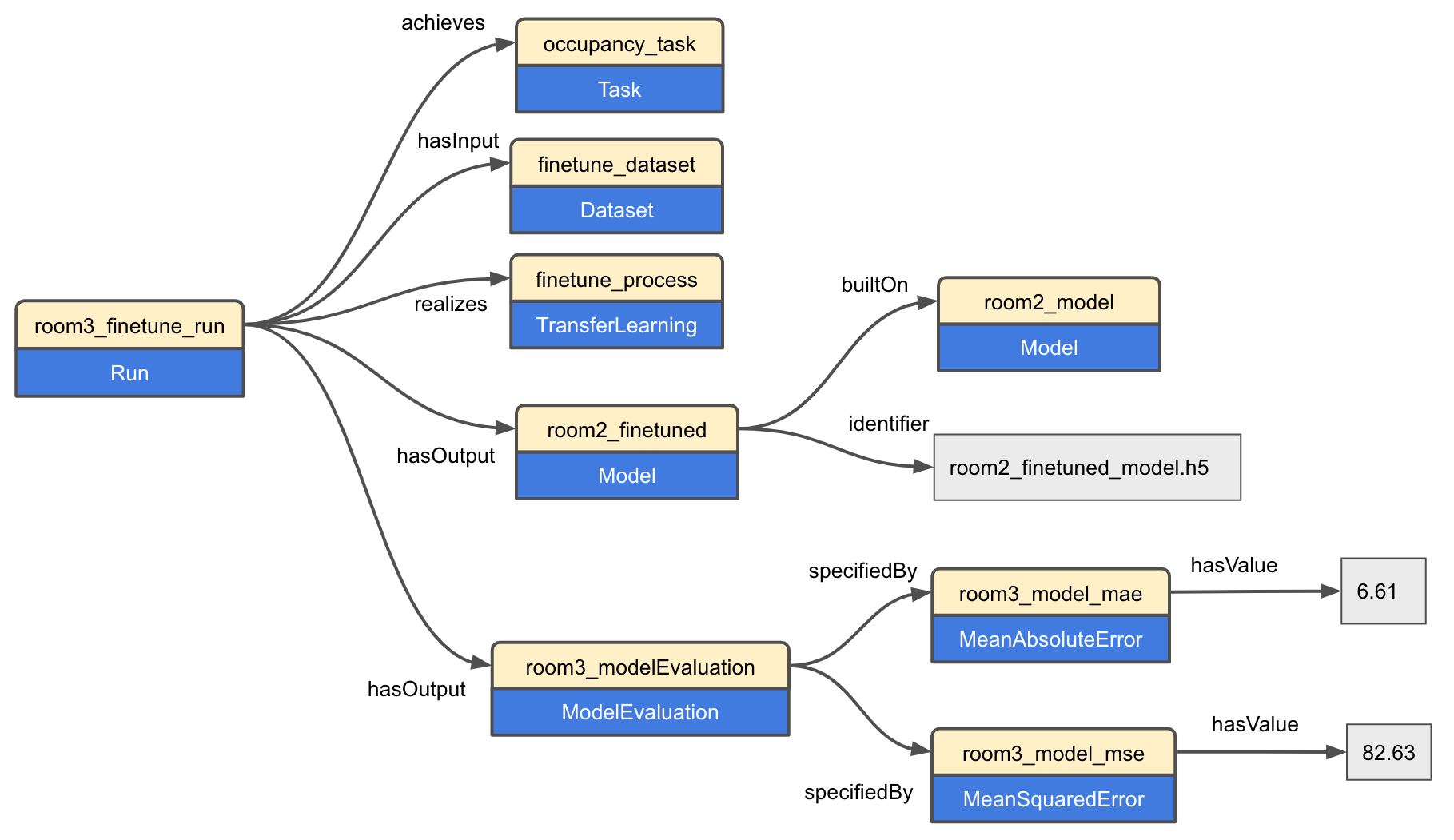}
\caption{Definition of a model in Knowledge Graph}
\label{knowledge_graph}
\end{figure}

As depicted in Figure \ref{knowledge_graph}, machine learning processes can be characterized based on their input data, output model, and model evaluation measures such as Mean Absolute Error (MAE) or Mean Squared Error (MSE) of learning processes. Furthermore, each model is dedicated to a specific task defined by ML-Schema which helps spaces to find the relevant models shared by other spaces. As such, the models are presented in a self-explainable and interpretable way to both human users as well as space agents (machines).

The training process depicted in Figure \ref{knowledge_graph} undertakes a transfer learning process. In this specific case, the process includes a fine-tuning process which takes the trained model of another room (model of Room-2) as input and retrain it based on a small fine-tuning dataset of target room (Room 3). 

\begin{figure}
\centering
\includegraphics[width=1\textwidth]{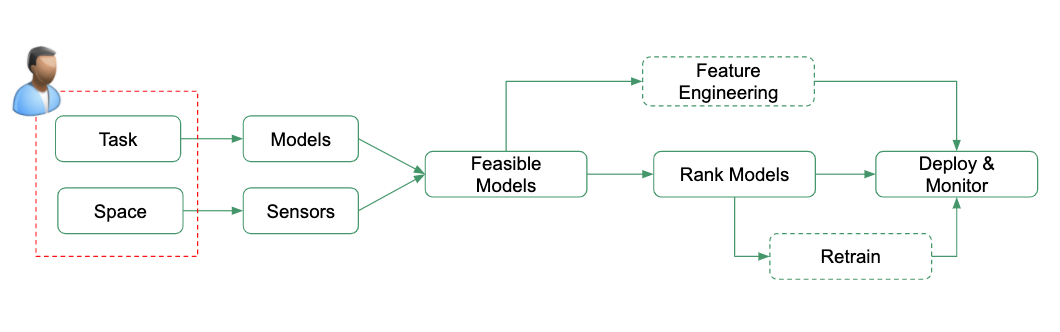}
\caption{Model query and reuse process.}
\label{model_query}
\end{figure}

The ISL frameworks envisions communication between multiple ISL nodes where each node represents the underlying knowledge of corresponding smart environment captured in the knowledge graph using a common schema of building assets (i.e. semantics of sensors, services, etc.). Based on this configuration users will be able to query services that match the characteristics of a given space (i.e., IoT devices such as sensors and actuators). As shown in Figure \ref{model_query}, in a model query scenario, user can match the IoT specification of target space with the offered models and retrieve the feasible models. For instance, in case of room occupancy detection based on CO2 sensors, the target room should be equipped with a CO2 sensor and all retrieved models should also accept CO2 data as input. In the next step, the retrieved models can be ranked based on quality indicators which are also available in knowledge graph and are offered by the ISL framework. As an extra step and if some training data is available in the target space, the model can be fine-tuned for a better performance. 

\section{Proposed Approach}

In order to realize this requirements, the ISL resources needs to be shared and advertised on blockchain by ISL nodes. These resources are usually not deterministic which contradicts with the nature of blockchain. The technical solution to address this issue is to use an Oracle Smart Contract which is responsible to retrieve and verify external data for blockchain and smart contracts through methods such as web APIs or data feeds.

In our case, we would need an oracle to interact with external ISL nodes that are willing to share and reuse machine learning models and their corresponding assets. In our implementation, the oracle smart contract is created by the network owner and manages list of trustworthy ISL nodes. It also allows the trustworthy ISL nodes to add their resources and maintains a registry of shared resources. 

In this section, we formally define the knowledge sharing framework and outline how we use the blockchain technology to facilitate the communication and knowledge exchange between ISL nodes.

Let $\mathcal{N}$ be the set of ISL nodes representing the digital twins of smart environments that are supposed to take part in transfer learning use cases. Each ISL node $\alpha \in \mathcal{N}$ is then characterized by its knowledge graph that we denote by $\mathcal{G}_\alpha$ and captures the description of machine learning assets such as training datasets and ML models. We define a local dataset as a dataset that is captured in $\mathcal{G}_\alpha$ and owned by the corresponding ISL node. 

We denote the collection of local datasets introduced in knowledge graph $\mathcal{G}_\alpha$ by  $\mathcal{D}|_{G_\alpha}$. Additionally, the subset of local datasets that are shared by the ISL node $\alpha$ are denoted by $\mathcal{D}^|_{G_\alpha}$ and we refer to them as blockchain-enabled datasets. Further, we denote the set of all blockchain-enabled datasets as $\mathcal{D}^*$ and define it as follows: 

\[
\mathcal{D}^* = \{\mathcal{D}^*|_{G_n} : n \in \mathcal{N} \}
\]

Consequently, the models that are introduced in $\mathcal{G}_\alpha$ can be trained using the available local or blockchain-enabled datasets to achieve a specific task. The set of all models introduced by the knowledge graph $\mathcal{G}_\alpha$ is defined as: 
\[
\mathcal{M}|_{G_\alpha} = \{ M_i \langle task, dataset, model\_uri, base\_model \rangle : i = 1,2,\ldots , m \} 
\]

Where task is an instance of class Task defined in the knowledge graph; model\_uri is the identifier of the target model file; and dataset is selected from $\mathcal{D}|_{G_\alpha} \cup  \mathcal{D}^*$. Furthermore, in case of models that are trained from scratch, no base model is needed and we set the base-model to null. However, in a transfer learning case, the new model reuses an existing base model that either resides on the same ISL node as the new model, or comes from the pool of shared models from other ISL nodes. Similar to shared datasets, we denote the subset of local models that are shared by the ISL node $\alpha$ as $\mathcal{M}^*|_{G_\alpha}$. We also denote the set of all blockchain-enabled models as $\mathcal{M}^*$ and define it as follows:

\[
\mathcal{M}^* = \{\mathcal{M}^*|_{G_n} : n \in \mathcal{N} \}
\]

So, the base\_model in a transfer learning case is selected from $\mathcal{M}|_{G_\alpha} \cup \mathcal{M}^*$. As such, each model $\mathcal{M}_n$ depends on its training/fine-tuning dataset $\mathcal{D}_n$ and its base\_model (if any) $\mathcal{M}_{n-1}$. Then the dependency chain of model $\mathcal{M}_n$ that include a number of blockchain-enabled models and datasets, can be constructed as follows:  
\[
\varnothing \xrightarrow{\mathcal{D}_0}\mathcal{M}_0 \xrightarrow{\mathcal{D}_1}\mathcal{M}_1 \rightarrow{\ldots} \xrightarrow{\mathcal{D}_n}\mathcal{M}_n
\]

\begin{figure}
\centering
\includegraphics[width=0.7\textwidth]{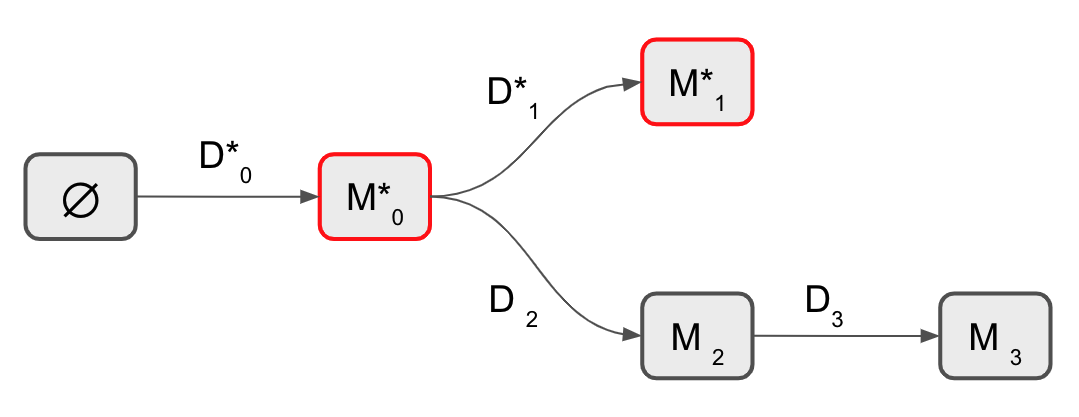}
\caption{Example of model dependency graph.}
\label{dependency_graph}
\end{figure}

At the root of all dependency chains, we have a model created based on a null model and using a specific dataset. This indicates that this model is created from scratch trained based on the specified dataset. Following this presentation, the collection of all dependency chains can be captured as a directed acyclic graph where vertices represent models and edges represent the corresponding dataset. Furthermore, all vertices, except for the root vertex, have in-degree one. An example of such dependency graph is depicted in Figure \ref{dependency_graph} where shared models are shown as red boxes. It's important to note that all assets of a shared model including ancestor models and the datasets used for training/fine-tuning in the dependency chain are required to be shard as well. In this way, one can easily trace back all the used assets of each model and investigate the authenticity of models and datasets.

\begin{figure}
\centering
\includegraphics[width=1\textwidth]{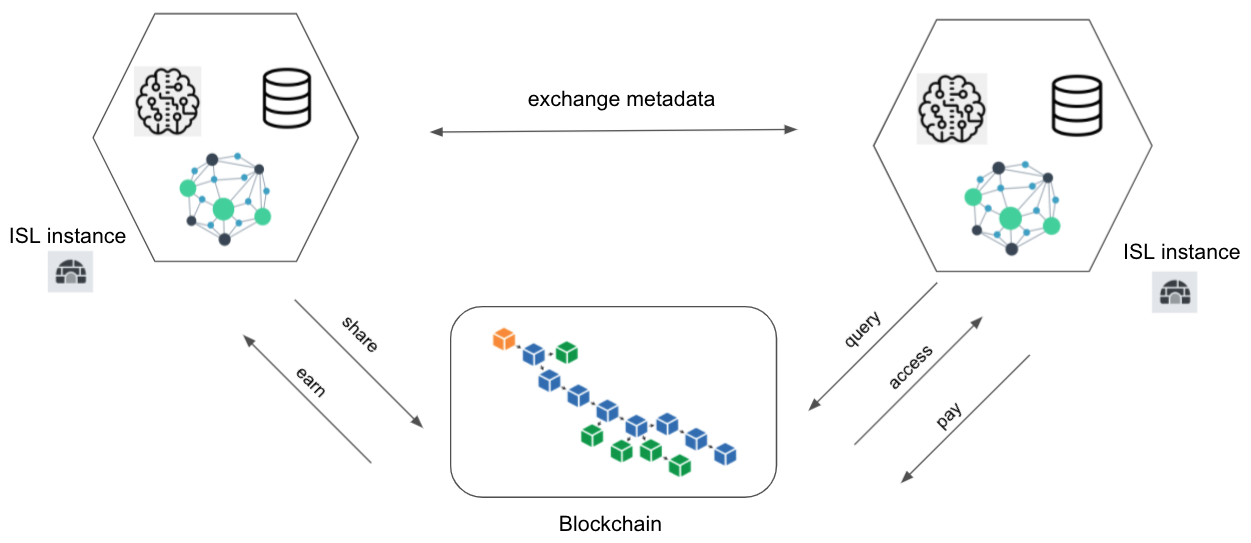}
\caption{Process of sharing and consuming of cognitive resources between smart dataspaces.}
\label{share_consume_process}
\end{figure}

As depicted in Figure \ref{share_consume_process}, the blockchain is used as shared medium between service providers and service consumers on ISL network to store the metadata of machine learning assets. 

Our proposed blockchain solution, includes an Oracle smart contract that keeps track of sharing resources and their dependency chains. More specifically, the Oracle at its basic level is defined as follows: 

\[
Oracle = \langle \mathcal{D}^*, \mathcal{M}^*,  task\_index\rangle
\]

Where task\_index maps task instances to models and facilitates searching for appropriate models.

\begin{figure}
\centering
\includegraphics[width=0.5\textwidth]{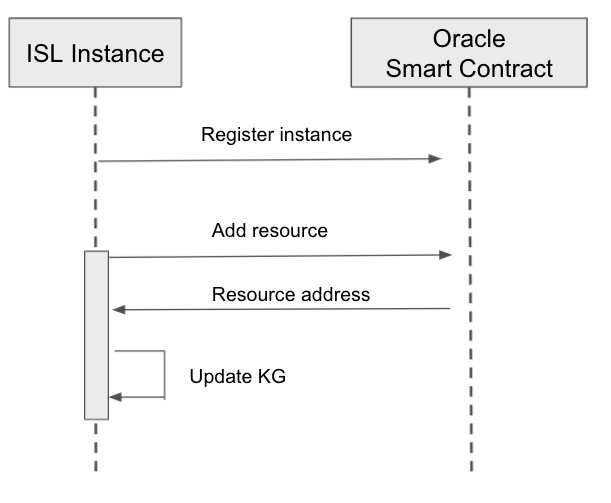}
\caption{Sharing resources on blockchain}
\label{share_process}
\end{figure}

The resource providers register their models and dataset via the Oracle smart contract as shown in Figure \ref{share_process}. To this end, the network owner needs to introduce the ISL node to blockchain by registering them to the oracle smart contract. Then, registered ISL nodes will be able to add the metadata of their resources to the oracle smart contract. In return, they receive a unique identifier that will be added to knowledge graph for further queries.

\begin{figure}
\centering
\includegraphics[width=0.5\textwidth]{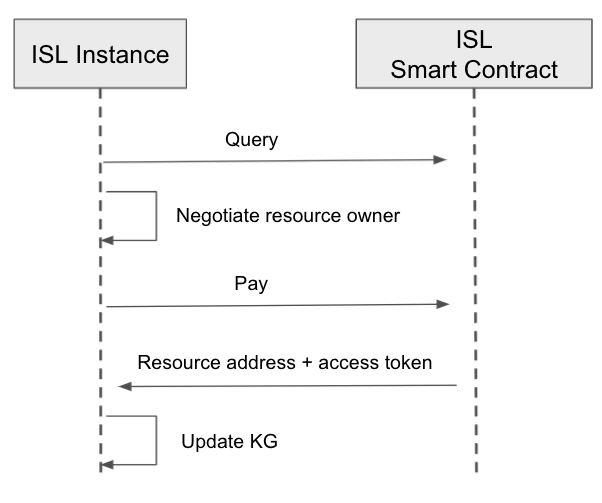}
\caption{Consuming blockchain resources.}
\label{consume_process}
\end{figure}

On the other hand, as shown in Figure \ref{consume_process}, the resource consumers can query the Oracle smart contract for the machine learning models that fulfill a specific task. Then, the consumer can negotiate the corresponding resource owners (ISL nodes) for finding the qualified resources. In the next step, the consumer will pay for the resource and receives the resource address and access token to retrieve the requested resource.

\section{Implementation}

\begin{figure}
\centering
\includegraphics[width=1\textwidth]{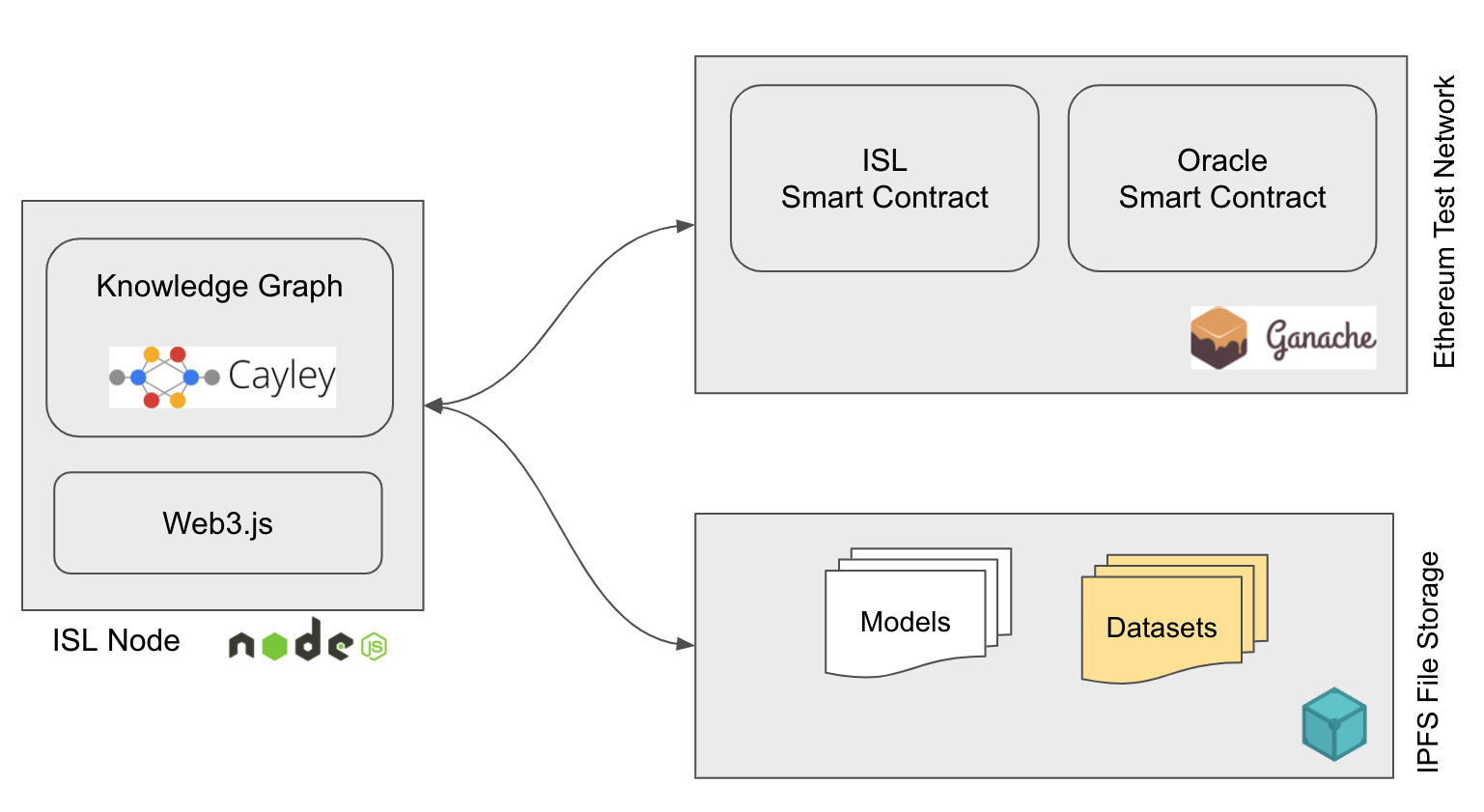}
\caption{Architecture of implemented prototype.}
\label{prototype_architecture}
\end{figure}

As proof of concept, we have implemented a prototype based on the proposed methodology. In this section, we provide an overview of the technical architecture which is depicted in Figure \ref{prototype_architecture} extends our previous implementation of ISL framework for knowledge sharing between smart environments \cite{anjomshoaa2021transfer}. 

At the core of ISL framework, we have the knowledge graphs of ISL nodes that include the description of spaces, IoT configurations, and machine learning assets. The knowledge graphs are maintained by each ISL node and stored using Cayley\footnote{https://github.com/cayleygraph}, an open-source database for Linked Data. The node implementation offers APIs for communication between ISL nodes as well as a web interface for displaying and visualizing the knowledge graph assets and entities. In the new implementation, we leverage the existing knowledge graphs in order to store the relevant information of blockchain environment and blockchain-enabled assets. This includes the information of blockchain accounts as well as the blockchain addresses of shared resources such as datasets and models.

In order to make the ISL nodes blockchain-enabled, the ISL nodes web application is upgraded to a decentralized application(dApp) implemented on Ethereum test network. We have used Ganache \footnote{https://www.trufflesuite.com/ganache} environment to develop and test the required smart contracts. To this end, each ISL nodes is equipped with a unique account address that is used to identify the corresponding ISL node on the blockchain. The implementation includes the following two smart contracts that are implemented in Solidity programming language \cite{dannen2017introducing}:

\begin{itemize}
  \item ISL smart contract which provides an interface for interaction between ISL nodes and organizes transactions such as acquiring resources (e.g.; datasets and models) and performing payment transactions if required.
  
  \item Oracle smart contract which is responsible for governance of participating ISL nodes and manages dedicated lists of shared resources. This smart contract also includes the account number of the oracle owner who has the authority to register new ISL nodes. As such, only trusted ISL nodes will be allowed to contribute their resources or acquire the shared resources from other trusted ISL nodes.
 \end{itemize}

In this architecture, each ISL node stores its resources off-chain in a  local file repositories which is only accessible to the ISL node itself. However, when the resource is shared on blockchain, we would need to provide access to the resource files to external ISL nodes. To this end, we use InterPlanetary File System (IPFS) which is a protocol and peer-to-peer network for storing and sharing data in a distributed file system. In our case, whenever an ISL node shares a resource such as dataset of model on blockchain, the underlying files will be stored on IPFS. Then, the unique content-address which is assigned by IPFS will be added to both the local knowledge graph and the corresponding list in oracle smart contract. Since the IPFS provides a universally unique address for each resource, the resources cannot be shared twice. As such, the ownership rights of resources can be checked and preserved by the oracle smart contract. 

\begin{figure}
\centering
\includegraphics[width=1\textwidth]{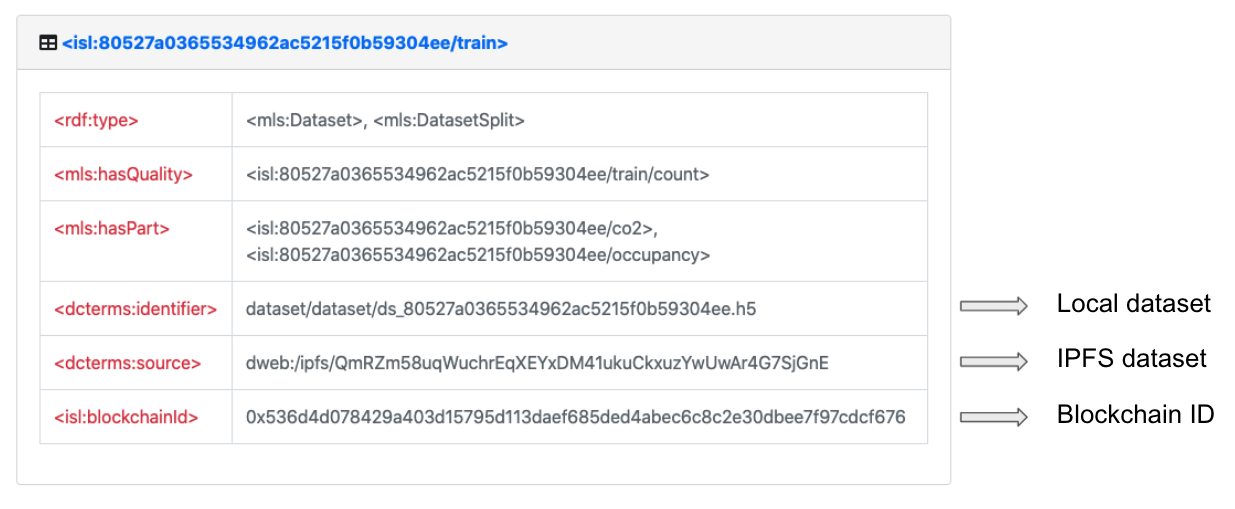}
\caption{Structure of blockchain-enabled resources in knowledge graph.}
\label{resource_structure}
\end{figure}

Figure \ref{resource_structure} demonstrates the representation of a blockchain-enabled dataset in the knowledge graph which includes the URI of the local dataset file, the URI of shared dataset on IPFS, and the unique transaction identifier which is assigned by the oracle smart contract.

\section{Related Work}
There is a handful of methodologies and platforms that facilitate knowledge management and data sharing for machine learning scenarios. In this section, we explore a number of existing solution and research work that are expanded on the fields of machine learning, blockchain, and data sharing frameworks. 

\subsection{Blockchain-based Machine Learning Approaches}

Machine learning solutions are commonly designed to be used within organization boundaries. However due to the costly approached for creating elaborated models, it makes sense to share and reuse the machine learning assets beyond organization borders. In this context we would need a trustworthy and sustainable approach to make the machine learning models accessible to external entities. 

Recently, several research works have used blockchain as share and reuse medium in machine learning use cases. For instance, DInEMMo \cite{marathe2018dinemmo} proposes a solution that combines machine learning and blockchain frameworks to create a marketplace for machine learning models that enables to reward both model creators as well as actors that contribute to enhance existing models. In \cite{Rathore_2019}, authors present BlockDeepNet, a Blockchain-based secure deep learning solution that combines deep learning and blockchain technologies to support secure collaborative learning in IoT environments. the work in \cite{justin2019decentralized} proposes a framework for participants to collaboratively build a dataset and use smart contracts to host a continuously updated model. This work also presents financial and non-financial (gamified) incentive structures for providing good data and includes a prototype implemented on Ethereum blockchain.

Another study on secure and resilient machine learning adoption for making blockchain-based smart applications has been presented in \cite{tanwar2019machine} where traditional machine learning  techniques are applied to analyze the attacks on a blockchain-based network. The study also discusses how these technologies can be applied in applications such as Unmanned Aerial Vehicle (UAV), Smart Grid (SG), healthcare, and smart cities. The work in \cite{xiong2019smart} proposes a solution for data trading based on smart contract using blockchain to protect the rights and interests of the data owner, and off-chain machine learning methods for addressing the data availability controversies. 

The work in \cite{kurtulmus2018trustless} have proposed a protocol based on the anonymous and distributed nature of smart contracts, and the problem-solving aspect of machine learning which helps users obtain machine learning models for a given fee. The protocol does not require trust and works completely on a decentralized blockchain. The work in \cite{harris2019decentralized} presents a configurable framework for training a model and collecting data on blockchain.

Decentralized machine learning protocol (DML) \cite{dml} project focuses on on-device machine learning, blockchain and federated learning technologies. It unleashes untapped data usage without extraction and idle processing power of user computers for machine learning. Algorithms will be crowdsourced from a developer community through the marketplace resulting innovation from periphery.
The GNY’s decentralized machine learning solution \cite{gny} aims to bring a secure, collaborative platform for machine learning, data capture and analysis to the Lisk blockchain project.

\subsection{Artifact Sharing in Machine Learning Environments}

Machine learning solution consist of multiple sequential steps that range from data processing and feature engineering to model training and deployment. There is a handful of existing solutions and research work for managing the life cycle of machine learning processes and pipelines.

ModelDB \cite{vartak2016modeldb} is an open source model management system that offers services for saving models and their associated metadata. It also includes services for measuring the performance of machine learning models and querying models. ModelHUB \cite{miao2017modelhub} is another platform for managing the lifecycle of deep learning models that provides services for model storage and keeps track of associated metadata and accuracy scores. Although these solutions facilitate the communication between team members and support pipelines for creating and managing machine learning workflows, however such solutions does not cross the organization borders and it is not easy to share the artifacts in a reliable and trustworthy way. 

Another project in this context is TensorFlow Extended (TFX) \cite{baylor2017tfx} which a solution built around is the Google's TensorFlow, an end-to-end open source platform for machine learning. By integrating various machine learning artifacts into one platform, it provides a standardize way to manage and access machine learning assets and artifacts which facilitates the platform configuration, and reduce the time to production significantly, while providing platform stability.

Michelangelo \cite{hermann2017meet} platform is the Uber's solution for building, deploying, and operating machine learning workflows in a collaborative environment and covers areas such as data management, model training, and model deployment, evaluation, and monitoring. An integral part of the Michelangelo platform is a sub-system called Gallery \cite{sun2020gallery} that manages the lifecycle of machine learning models by orchestrating the flow of models across different stages in the lifecycle. 

There are also other solutions that are more focused on data pipelines and feature management aspects. For instance, Hopsworks Feature Store \cite{kakantousis2019horizontally} provides an open data management ecosystem for machine learning features, including the feature engineering code and the feature data.

\section{Conclusions and Future Work}
Increasing access to machine learning artifacts held by various organizations will accelerate the uptake of use cases and solutions for smart environments such as smart home and smart buildings. Due to increasing number of IoT-enabled environments and the costs associated with machine learning processes, sharing and reuse  of machine learning assets and artifacts beyond organization borders are plausible solutions. In this paper, we proposed a decentralized and adaptive software framework based on blockchain and knowledge graph technologies that supports the knowledge exchange and interoperability between smart environments, in a transparent and trustworthy way. The proposed architecture can be used in ways that increase communication between smart environments, enable sustainable growth of IoT solutions, and drive sharing of cognitive knowledge via transfer learning. In this context, self-explaining and machine-understandable digital assets (e.g., features and models) are the key requirements to enable transfer learning beyond the organizational borders. Also, blockchain technology and smart contracts foster transparent and trustworthy sharing of machine learning assets and advance cross-organization communication between smart environments.

\section{Acknowledgement}
This work was supported with the financial support of the Science Foundation Ireland grant 13/RC/2094 and co-funded under the European Regional Development Fund through the Southern \& Eastern Regional Operational Programme to Lero - the Irish Software Research Centre (www.lero.ie).

\bibliographystyle{splncs04}
\bibliography{bibliography}

\begin{thebibliography}{10}
\providecommand{\url}[1]{\texttt{#1}}
\providecommand{\urlprefix}{URL }
\providecommand{\doi}[1]{https://doi.org/#1}

\bibitem{anjomshoaa2021transfer}
Anjomshoaa, A., Curry, E.: Transfer learning in smart environments. Machine
  Learning and Knowledge Extraction  \textbf{3}(2),  318--332 (2021)

\bibitem{baylor2017tfx}
Baylor, D., Breck, E., Cheng, H.T., Fiedel, N., Foo, C.Y., Haque, Z., Haykal,
  S., Ispir, M., Jain, V., Koc, L., et~al.: Tfx: A tensorflow-based
  production-scale machine learning platform. In: Proceedings of the 23rd ACM
  SIGKDD International Conference on Knowledge Discovery and Data Mining. pp.
  1387--1395 (2017)

\bibitem{dannen2017introducing}
Dannen, C.: Introducing Ethereum and solidity, vol.~318. Springer (2017)

\bibitem{dml}
DML, D.: Decentralized machine learning protocol (dml),
  \url{https://decentralizedml.com/}, [Online; accessed 15-July-2021]

\bibitem{gny}
GNY: Gny decentralized machine learning), \url{https://www.gny.io/}, [Online;
  accessed 15-July-2021]

\bibitem{harris2019decentralized}
Harris, J.D., Waggoner, B.: Decentralized and collaborative ai on blockchain.
  In: 2019 IEEE International Conference on Blockchain (Blockchain). pp.
  368--375. IEEE (2019)

\bibitem{hermann2017meet}
Hermann, J., Del~Balso, M.: Meet michelangelo: Uber’s machine learning
  platform. URL https://eng. uber. com/michelangelo  (2017)

\bibitem{justin2019decentralized}
Justin, D., Harris, B.: Decentralized \& collaborative ai on blockchain. In:
  Proceedings of the 2019 IEEE International Conference on Blockchain
  (Blockchain), Atlanta, GA, USA. pp. 14--17 (2019)

\bibitem{kakantousis2019horizontally}
Kakantousis, T., Kouzoupis, A., Buso, F., Berthou, G., Dowling, J., Haridi, S.:
  Horizontally scalable ml pipelines with a feature store. In: Proc. 2nd SysML
  Conf., Palo Alto, USA (2019)

\bibitem{kurtulmus2018trustless}
Kurtulmus, A.B., Daniel, K.: Trustless machine learning contracts; evaluating
  and exchanging machine learning models on the ethereum blockchain. arXiv
  preprint arXiv:1802.10185  (2018)

\bibitem{marathe2018dinemmo}
Marathe, A., Narayanan, K., Gupta, A., Manoj, P.: Dinemmo: Decentralized
  incentivization for enterprise marketplace models. In: 2018 IEEE 25th
  International Conference on High Performance Computing Workshops (HiPCW). pp.
  95--100. IEEE (2018)

\bibitem{miao2017modelhub}
Miao, H., Li, A., Davis, L.S., Deshpande, A.: Modelhub: Deep learning lifecycle
  management. In: 2017 IEEE 33rd International Conference on Data Engineering
  (ICDE). pp. 1393--1394. IEEE (2017)

\bibitem{Rathore_2019}
Rathore, S., Pan, Y., Park, J.H.: Blockdeepnet: A blockchain-based secure deep
  learning for iot network. Sustainability  \textbf{11}(14), ~3974 (Jul 2019).
  \doi{10.3390/su11143974}, \url{http://dx.doi.org/10.3390/su11143974}

\bibitem{sun2020gallery}
Sun, C., Azari, N., Turakhia, C.: Gallery: A machine learning model management
  system at uber. In: EDBT. pp. 474--485 (2020)

\bibitem{tanwar2019machine}
Tanwar, S., Bhatia, Q., Patel, P., Kumari, A., Singh, P.K., Hong, W.C.: Machine
  learning adoption in blockchain-based smart applications: The challenges, and
  a way forward. IEEE Access  \textbf{8},  474--488 (2019)

\bibitem{vartak2016modeldb}
Vartak, M., Subramanyam, H., Lee, W.E., Viswanathan, S., Husnoo, S., Madden,
  S., Zaharia, M.: Modeldb: a system for machine learning model management. In:
  Proceedings of the Workshop on Human-In-the-Loop Data Analytics. pp.~1--3
  (2016)

\bibitem{xiong2019smart}
Xiong, W., Xiong, L.: Smart contract based data trading mode using blockchain
  and machine learning. IEEE Access  \textbf{7},  102331--102344 (2019)

\end{thebibliography}

\end{document}